\pdfoutput=1

\documentclass[11pt]{article}

\usepackage[]{acl}

\usepackage{times}
\usepackage{latexsym}

\usepackage[T1]{fontenc}

\usepackage[utf8]{inputenc}

\usepackage{microtype}

\usepackage{multirow}
\usepackage{graphicx}
\usepackage[fleqn]{amsmath}

\usepackage{tabularx}
\usepackage{siunitx}
\newcommand{\genname}{Diversity Threshold Generation}
\newcommand{\gennamespace}{Diversity Threshold Generation }

\newcommand{\nli}{NLI Diversity}
\newcommand{\nlispace}{NLI Diversity }

\usepackage{enumitem}
\setlist[itemize]{noitemsep}
\usepackage{booktabs}

\newcommand\Tstrut{\rule{0pt}{2.6ex}}

\title{Semantic Diversity in Dialogue with Natural Language Inference}

\author{Katherine Stasaski and Marti A. Hearst \\
  UC Berkeley\\
  \texttt{\{katie\_stasaski, hearst\}@berkeley.edu} \\}

\begin{document}
\maketitle
\begin{abstract}
Generating diverse, interesting responses to chitchat conversations is a problem for neural conversational agents. 
This paper makes two substantial contributions to improving diversity in dialogue generation.  First, we propose a novel metric which uses  Natural Language Inference (NLI)  to measure the semantic diversity of a \textit{set} of model responses for a conversation.  
We evaluate this metric using an established  framework \cite{tevet-berant-2021-evaluating} and find strong evidence indicating \nlispace is correlated with semantic diversity.
Specifically, we show that the contradiction relation is more useful than the neutral relation for measuring this diversity and that incorporating the NLI model's confidence achieves state-of-the-art results. 
Second,
we demonstrate how to iteratively improve the semantic diversity of a sampled set of responses via a new generation procedure called \genname,
which results in an average 137\%  increase in \nlispace compared to standard generation procedures.  
\end{abstract}

\section{Introduction}
 
Dialogue models often struggle to produce engaging utterances in conversations, tending to generate responses which are common in the training data, such as ``OK,'' ``Yeah,'' or ``I don't know'' \cite{li-etal-2016-diversity}.  While these responses are appropriate for a wide variety of contexts, their over-production can result in a dull conversation \cite{see-etal-2019-makes}.

An evaluation task has emerged that consists of measuring the diversity of chitchat model responses over a test set.  While some past work uses human evaluation to measure model response diversity according to engagingness, specificity, or interestingness  \cite{li-etal-2016-diversity, see-etal-2019-makes, Ghandeharioun-etal}, several automated metrics have also been proposed to measure diversity of model responses.  Some metrics measure \textit{lexical} diversity, typically via n-gram overlap  \cite{li-etal-2016-diversity} or computing the BLEU score \cite{zhu-et-al-acm} among model responses generated from the test set.  Other past work attempts to measure \textit{semantic} diversity via repurposing sentence similarity metrics
\cite{tevet-berant-2021-evaluating, ZhangKWWA20, cer-etal-2017-semeval}.

\begin{figure}[t]
\includegraphics[width=7cm]{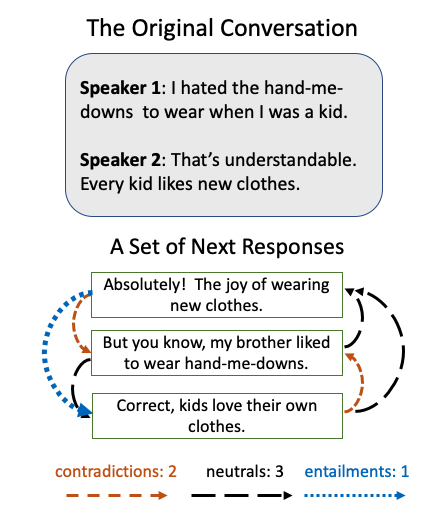}
\caption{Illustration of \nlispace using human responses from DailyDialog++. Contradictions are weighted by 1, entailments by -1, and neutrals by 0, so
the score is $(2\times1) + (3\times0) +(1\times-1) = 1$.  
}
\label{fig:process_fig}
\end{figure}

We propose a new metric aimed at measuring semantic diversity by leveraging a Natural Language Inference (NLI) model to score a set of multiple dialogue model responses for a single conversation, as illustrated in Figure \ref{fig:process_fig}.  NLI is a three-way classification task to determine whether one sentence entails, contradicts, or is neutral toward a second sentence.
We hypothesize that a diverse set of responses for a conversation captures contradictory ways one could respond, which can be measured by the NLI model. 
We aggregate the contradiction, neutral, and entailment predictions among pairs of responses from the set and combine the 
predictions into a new diversity metric, called \textit{\nli}.  

We additionally explore two modifications of \nli.
First, because the neutral prediction may be indicative of diversity, we propose Neutral \nli, where neutral predictions are weighted the same as contradiction predictions. Second, since our Baseline \nlispace method does not take into account the \textit{confidence} of the model's prediction, we propose Confidence \nli, which aggregates the probability mass of the model's predicted class instead of aggregating the number of predictions for each class.

We assess \nlispace using \citet{tevet-berant-2021-evaluating}'s diversity metric evaluation framework, finding that \nlispace is highly correlated both with human judgments of diversity and with the \textit{diversity parameter}, a gold standard diversity value used to generate the set of responses.
Confidence \nlispace achieves state-of-the-art performance in terms of correlation with semantic diversity.  
Also, through an ablation study, we find  positive, neutral, and negative correlations between human judgments and the number of contradiction, neutral, and entailment predictions, respectively.

We next explore 
the use of a dialogue model to generate a set of candidate responses with a minimum target
level of semantic diversity, such as 10 Contradictions.  Our new generation procedure, \textit{\genname}, iteratively improves a set of model responses until this intended threshold is reached.
If a set of sampled responses does not meet the intended threshold, the lowest-scoring response is thrown out and a new response is sampled until the diversity threshold is reached.  
We show this procedure results in a more diverse set of responses than the original sampled set, often with only a few resampled responses. Results of automated analysis shows relevancy is maintained from initial to final sets of responses.

In summary, our contributions are:
\begin{itemize}[noitemsep,topsep=0pt]
    \item A novel 
    diversity metric, \nli, evaluated using \citet{tevet-berant-2021-evaluating}'s framework, that measures semantic diversity and interrogates the relationship between Contradiction and Neutral predictions and diversity,
    \item Confidence \nli, a diversity metric which obtains state-of-the-art performance on semantic diversity,
    \item A new dialogue generation procedure, \genname, which continues sampling  responses until an intended diversity threshold, defined using \nli, is reached,
    \item Experimental results indicating dialogue models are able to generate diverse responses using \gennamespace with minimal loss in relevancy.
\end{itemize}
\section{Related Work}

Past work has explored lexical and semantic diversity metrics as well as ways of evaluating these metrics.  We also draw from work in NLI and generating diverse sets of hypotheses.

\subsection{Measuring Model Response Diversity}
Traditionally, a model's diversity has been measured in terms of its predictions over the test set \cite{li-etal-2016-diversity}, which we call \textit{Test Set Diversity}.  In this setup, the model predicts one response for each conversation in the test set (containing $n$ conversations), resulting in $n$ predictions.  The diversity measure is computed over these $n$ predictions, resulting in a 
score over the entire test set.

The notion of diversity we investigate, however, measures the model's ability to generate a \textit{set} of responses for a single conversation \cite{zhang-etal-2019-syntax-infused, tevet-berant-2021-evaluating}, which we call \textit{Multi-Response Diversity}.  Instead of generating one response for each of the conversations in the test set, we evaluate a model's ability to generate $m$ responses for each of the $n$ conversations.  

As shown by \citet{tevet-berant-2021-evaluating}, metrics which have been proposed in the \textit{Test Set Diversity} setting can still be applied in the \textit{Multi-Response Diversity} setting, however,
by treating each set of $m$ responses as its own ``test set'' and averaging over the $n$ total sets.

\subsection{Diversity Metrics}
Lexical diversity metrics measure differences in word choice, as opposed to diversity of content.  \citet{li-etal-2016-diversity} propose \textit{distinct-n}, which measures the number of unique n-grams generated 
divided by the total number of n-grams generated
in the \textit{Test Set Diversity} setting.
Some past work has applied this metric to the \textit{Multi-Response Diversity} setting 
\cite{tevet-berant-2021-evaluating}.  
\citet{cao-clark-2017-latent} propose examining the percent of unique \textit{responses}  over the test set.
Other past work has proposed using BLEU score over a set of model responses in the \textit{Test Set Diversity} setting \cite{zhu-et-al-acm}.

Semantic diversity metrics, on the other hand, compare diversity of the content present in each response.  Many of these measures are adapted from semantic similarity scores, since lower similarity can indicate higher diversity \cite{tevet-berant-2021-evaluating}.  BERTScore measures the similarity of BERT embeddings for each token in two sentences
\cite{ZhangKWWA20}.  Bert-STS assigns a score 
based on the semantic similarity of two sentences \cite{tevet-berant-2021-evaluating}.  The Sent-BERT metric computes cosine similarity between BERT sentence embeddings \cite{reimers-gurevych-2019-sentence}.  \citet{larson-etal-2019-outlier} propose identifying diverse paraphrases 
by identifying embedding outliers.

Other past work has used human evaluation to measure a model's diversity.  \citet{li-etal-2016-diversity} ask humans to choose the better of two responses 
based on specificity to the past conversation.  \citet{see-etal-2019-makes} ask humans to rank dialogue responses on a variety of factors, including interestingness and inquisitiveness.  \citet{tevet-berant-2021-evaluating} compare participants' ability to judge diversity of a set of responses in two ways: (i) by  ranking one response as more diverse than a second response and (ii) by judging the diversity of a single response on a Likert scale, finding that participants were equally able to judge diversity in both conditions.  They also find that human judges are better at distinguishing semantic diversity than lexical diversity.

Other past work has incorporated diversity metrics into the dialogue dataset creation pipeline.  \citet{stasaski-etal-2020-diverse} propose a method which measures the diversity of a crowdworker's contributions  compared to a corpus, using that information to determine when to stop collecting data from the worker.  This results in a more diverse  dataset. 

\subsection{Evaluation of Diversity Metrics}

\citet{tevet-berant-2021-evaluating} propose a framework to examine the reliability of diversity metrics.  They propose the notion of a \textit{diversity parameter}, which is used to generate a set of model responses, e.g.,
the $p$-value in nucleus sampling, which specifies the vocabulary probability distribution cutoff used to restrict sampling to the most-likely words whose combined likelihood $\geq p$.  
If $p$ is higher, the set of responses should have higher diversity, and vice-versa.   This \textit{diversity parameter} is treated as a gold standard for a set of responses' diversity.
Diversity metrics assign scores in the \textit{Multi-Response Diversity} condition and are evaluated in terms of correlation to the diversity parameter.  
They further propose two datasets  to evaluate diversity metrics: one which includes model responses and contains varying levels of lexical diversity 
and one which is human-created and maintains high lexical diversity to allow focused evaluation of semantic diversity.  

\subsection{Natural Language Inference}
Natural Language Inference is a task aimed at predicting whether one sentence contradicts, entails, or is neutral towards a second sentence. Models for NLI are typically trained using one of two datasets:  Stanford Natural Language Inference (SNLI) \cite{bowman-etal-2015-large} or Multi-Genre NLI (MNLI) \cite{mnli}.  
More recent datasets include FEVER \cite{thorne-etal-2018-fever, nie2019combining}, adapted from a fact-checking dataset, and ANLI \cite{nie-etal-2020-adversarial}, collected in an adversarial human-in-the-loop procedure.
With the rise of transformer architectures, models have achieved high performance on NLI tasks
\cite{liu2019roberta}.

In a dialogue setting, NLI has been used to improve \textit{consistency} between a 
persona and model responses over the course of a conversation by integrating an NLI-based reward into a reinforcement learning training procedure \cite{SongZH020}. 

To our knowledge, however, NLI has not been used to measure the diversity of model responses in either the \textit{Test Set Diversity} or the \textit{Multi-Response Diversity} setting.

\subsection{Generating Diverse Sets of Hypotheses}
While  work has only recently begun to explore the task of generating multiple \textit{dialogue} responses to a conversation \cite{zhang-etal-2019-syntax-infused, tevet-berant-2021-evaluating}, past work has explored generating diverse sets of hypotheses in some other application areas.  \citet{carbonell} explored using Maximal Mutual Relevance to reduce redundancy without sacrificing relevancy in document selection for summarization.  \citet{batra} proposed a greedy iterative algorithm to generate diverse, probable hypotheses for multiple vision tasks.  Most related to our work is \citet{gimpel-etal-2013-systematic}, which applied \citet{batra}'s approach to machine translation, generating a \textit{set} of translations instead of a single translation.  
In contrast to \citet{gimpel-etal-2013-systematic}, by holding the sampling procedure constant throughout the iterative process, our method can explore the extent to which diversity can be increased without altering standard decoding practices. 

\section{\nlispace Metric}\label{sec:nli}
We propose three diversity metrics in the \textit{Multi-Response Diversity} setting which leverage the predictions of an NLI model.  Two metrics (Baseline and Neutral) aggregate the NLI model's class predictions and one metric (Confidence) aggregates the weight of these predictions.

\subsection{Baseline \nli}
We propose a new metric, called \textit{Baseline \nli}, which uses an NLI model's predictions to measure diversity.
More formally, for a given conversation, $c$, and a dialogue generation model $M$, a set of utterances $u_1,...,u_n$ is produced by the model.  Each pair of  utterances is compared in both directions using an NLI model, $NLI(u_1,u_2),NLI(u_2, u_1),...,NLI(u_n,u_{n-1})$.  

The NLI model predicts a distribution over the three potential classes:  contradiction, neutral, and entailment.  We take the argmax over these classes,
resulting in a list of NLI predictions, $NLI_{preds}(NLI(u_1,u_2),...,NLI(u_{n-1},u_n))$ of size $n(n-1)$.  
To produce an overall diversity score for $NLI_{preds}(u_1,...,u_n)$, we assign each of these classes a value representing their diversity, denoted $NLI_{score}(NLI_{preds}(u_1,...,u_n))$.  

We hypothesize that larger numbers of entailment predictions found in a set of model-generated utterances is indicative of a lack of diversity; similarly, larger number of contradiction predictions is indicative of a larger amount of diversity.  
Because we want a higher value of $NLI_{score}$ to indicate higher diversity, we assign values as:
\[   
NLI_{score} = 
     \begin{cases}
       \text{1} & \text{if contradiction} \\
       \text{0} & \text{if neutral}\\
       \text{-1} &\text{if entailment} \\
     \end{cases}
\]
The sum of the $NLI_{score}$ values for the set of utterances results in the final \nlispace score, formally defined as:
\[
\begin{aligned}
    &Baseline~ \nlispace = \\
         &\sum_{u_i, u_j \in u_1,...,u_n} NLI_{score}(NLI_{pred}(NLI(u_i,u_j))
    \end{aligned}
\]

While the Baseline \nlispace metric aggregates all classes, we also investigate the separate number of entailment, contradiction, and neutral predictions in $NLI_{preds}$, denoted \# Entailment, \# Contradiction, and \# Neutral, respectively.

\subsection{Neutral \nli}
Our primary hypothesis is that contradictions indicate diversity and entailments indicate lack of diversity.  Because it is unclear what the role of neutrals might be, we  explore a version of \nlispace which weights neutral and contradiction predictions as equally diverse.  This metric is the same as Baseline \nlispace except the $NLI_{score}$ used to assign values is:
\[   
NLI_{score\_neutral} = 
     \begin{cases}
       \text{1} & \text{if contradiction} \\
       \text{1} & \text{if neutral}\\
       \text{-1} &\text{if entailment} \\
     \end{cases}
\]

\subsection{Confidence \nli}
Because the prior two \nlispace metrics do not incorporate the confidence of the NLI model's class predictions, we explore an additional metric which incorporates this value.  Letting $conf_{class}(u_1, u_2)$ represent the model's probability mass assigned to the predicted NLI $class$ after $softmax$, the function is defined as: $NLI_{score\_confidence} =$
\[
     \begin{cases}
       \text{1} \times conf_{con}(u_1, u_2) & \text{if contradiction} \\
       \text{0} & \text{if neutral}\\
       \text{-1} \times conf_{ent}(u_1, u_2) &\text{if entailment} \\
     \end{cases}
\]

\noindent
Intuitively, instead of assigning a 1 value for a contradiction prediction, this metric assigns the probability of the contradiction class.  Likewise, instead of a -1 for an entailment prediction, this metric assigns the negative probability mass of the entailment class.

\section{Evaluation of \nli}
We evaluate \nlispace by computing the correlation between the metric and both human labels and \textit{diversity parameter} labels.  Below we first describe the models and data and then present the results of the evaluation.

\subsection{Models}
We explore two NLI models:  a Roberta-large model \cite{liu2019roberta} fine-tuned on the Multi-Genre NLI (MNLI) Corpus \cite{mnli}\footnote{\url{https://huggingface.co/roberta-large-mnli}} and a Roberta-large model fine-tuned on a combination of MNLI, SNLI, FEVER, and ANLI\footnote{\url{https://huggingface.co/ynie/roberta-large-snli_mnli_fever_anli_R1_R2_R3-nli}}, both containing 300M parameters.  We refer to these models as \textit{\nlispace -- MNLI} and \textit{\nlispace -- Combined}, respectively.  We do not employ additional fine-tuning of these models.

\subsection{Data}

\begin{table}[t]
    \centering
    \renewcommand{\arraystretch}{1.2}
      \begin{tabular}{|p{2cm}  p{1.8in}|}
    \hline
        \textbf{decTest} &  
        Mixed Lexical Diversity; Mixed Semantic Diversity;  Model Generated   \\\cline{2-2}
   \textbf{Examples:}   & \\   
  temp 0.28              & ``I think he is the most awesome guy ever''\\
       & ``He is the most awesome guy ever''\\\cline{2-2}
temp 0.55  & ``The unemployment rate is lower than what it is''\\
   & ``No but it does make it more likely to be higher than what it is''\\
        \hline \hline
        \textbf{conTest} &  
        High Lexical Diversity; Mixed Semantic Diversity;  Human Generated  \\\cline{2-2}
 \textbf{Examples:}      & \\
 
 high lexical and     &   ``Sorry, but I don't agree.''\\
  low semantic   &      ``I think you are wrong about that.''\\\cline{2-2}
      & ``Dont be so judgemental, try to see \\
  high lexical and      & things her way.''\\
   high semantic      &  ``You are right that is insane.''\\
        \hline
\end{tabular}
\caption{Descriptions of diversity datasets from \citet{tevet-berant-2021-evaluating}. Corresponding temperature parameter (higher is more diverse) or semantic and lexical diversity levels accompany each 
example.}
\label{tab:div_data}
\end{table}

There are two different English datasets released to evaluate diversity metrics in \citet{tevet-berant-2021-evaluating}:  \textit{conTest} and \textit{decTest}, described in Table \ref{tab:div_data}.  The \textit{conTest} dataset is human-created and captures \textit{content}, or \textit{semantic}, diversity independent of \textit{lexical} diversity.  Low-diversity examples in this dataset have high lexical diversity but low semantic diversity.  This dataset was created by asking crowdworkers 
to generate sets of utterances with either low or high \textit{semantic} diversity using varied language, in order to keep a high level of \textit{lexical} diversity constant across both conditions.  

The \textit{decTest} dataset includes model-generated responses, with diversity  controlled by a decoding parameter, such as a temperature parameter.  The dataset can include duplicate responses, and does not attempt to mediate lexical diversity; therefore, low-diversity examples in this dataset may reflect \textit{low lexical} 
as well as \textit{low semantic} diversity.

While the original dataset 
includes multiple generation tasks, we evaluate on the dialogue task, \textit{respGen}, which is drawn  from Reddit conversations \cite{hashimoto-etal-2019-unifying}\footnote{In the data released from \citet{tevet-berant-2021-evaluating}, these files are called con\_test\_200\_with\_hds\_resp\_gen.csv and dec\_test\_200\_with\_hds\_resp\_gen.csv for \textit{conTest} and \textit{decTest}, respectively.}.  There are 200 conversations for each of \textit{conTest} and \textit{decTest} for the \textit{respGen} task, with multiple 
responses for each conversation (5 for \textit{conTest}, 10 for \textit{decTest}).

\subsection{Diversity Parameter Correlation}\label{sec:divparam}

The \textit{diversity parameter} from \citet{tevet-berant-2021-evaluating} represents either a parameter directly used to generate responses via a dialogue model, such as $p$ in nucleus sampling, or a binary value indicating whether crowdworkers were instructed to generate a high- or low-diversity set of responses.  A  measure which is able to capture diversity will be positively correlated with this diversity parameter.

\begin{table}[t]
\begin{center}
\def\arraystretch{1.2}
\begin{tabular}{ |p{4.3cm}|p{1.1cm}|p{1.1cm}| } 
 \hline
  & \textbf{decTest} & \textbf{conTest}\\
 \textbf{Metric} & $\rho$ & $\rho$ \\ \hline
Human Performance (absHDS) &	\textbf{0.81} & \textbf{0.63}	\\\hline

distinct-n	&	\textbf{0.89} & 0.34	\\
cos-sim	&	\textbf{0.89} &0.33	\\
BERT-STS	&	0.81 &0.46	\\
Sent-BERT	&	0.80 & 0.59	\\
BERTScore	&	0.87 &0.49	\\\hline\hline
Baseline \nlispace -- MNLI	&	0.58 & 0.59	\\
Baseline \nlispace -- Combined	&	0.39 & 0.59	\\
 \hline
Neutral \nlispace & 0.72 & 0.24 \\
Confidence \nlispace & 0.44 & \textbf{0.62}\\\hline
 
\end{tabular}
\end{center}

\caption{Spearman's $\rho$ correlations between \nlispace metrics and the diversity parameter.
Results above the double line are reproduced from \citet{tevet-berant-2021-evaluating}.
Both the best automatic metric and human performance for each dataset are in boldface.  
}
\label{tab:frameworkeval}
\end{table}

Table \ref{tab:frameworkeval} shows Spearman's correlations between \nlispace and the diversity parameter. 
On the \textit{conTest} semantic diversity dataset, Confidence \nlispace achieves the highest correlation of all metrics (0.62) 
and approaches human performance. Baseline \nlispace  performs comparably to the top-performing automatic metric from \citet{tevet-berant-2021-evaluating}, at 0.59 correlation.   
We note the 95\% confidence intervals overlaps between Baseline \nli, Confidence \nli, Sent-BERT, and human judgements, indicating a lack of significant differences (see Appendix \ref{app:ci}).
Although Neutral \nlispace does relatively poorly on \textit{conTest} (0.24), it is the highest-performing NLI metric on \textit{decTest} (0.72), suggesting that incorporating neutral predictions may capture lexical instead of semantic diversity.

\begin{figure}[t]
\includegraphics[width=7.5cm]{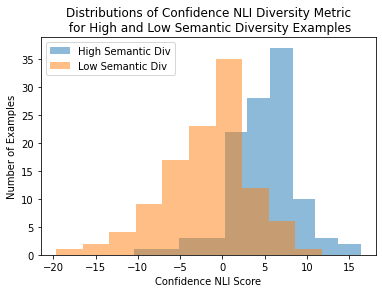}
\caption{Histogram of Confidence \nlispace for high and low semantic diversity examples.  }
\label{fig:dual_hist}
\end{figure}

A histogram of Confidence \nlispace values for low and high semantic diversity sets of responses is shown in Figure \ref{fig:dual_hist}.  We note the lack of large overlap between the distributions of low and high semantic diversity data.  In addition to the correlation results in Sections \ref{sec:divparam} and \ref{sec:humandiv}, this result indicates the Confidence \nlispace metric distinguishes between low and high semantic diversity.   

The higher correlation to the diversity parameter 
leads us to choose \nlispace - MNLI instead of Combined for all further experimentation.

\subsection{Human Correlation}\label{sec:humandiv}

\begin{table}
\begin{center}
\def\arraystretch{1.2}
\begin{tabular}{ |p{3.2cm}|S[table-format=1.2,table-column-width=1.5cm]|S[table-format=1.2,table-column-width=1.5cm]| } 
 \hline
{\parbox{1.5cm}{\textbf{Metric}}}	&	{\parbox{1.5cm}{\center \textbf{decTest
$\rho$}}}	&	{\parbox{1.5cm}{\center \textbf{conTest
$\rho$}}}	\Tstrut\\\hline
Baseline \nli	&	0.48	&	0.63\\
 Neutral \nli &  {\bfseries \num{0.69}} & 0.40\\
 Confidence \nli & 0.41 & {\bfseries \num{0.64}} \\\cline{1-3}
 \# Contradiction	&	0.26	&	0.46 \\
 \# Neutral	&	0.05	&	-0.08	\\
 \# Entailment	&	-0.48	&	-0.65	\\\hline

\end{tabular}
\end{center}

\caption{Spearman's $\rho$ correlation between
\nlispace metrics (MNLI) and human judgments.
 Negative values
indicate  higher $\#$ Entailment is \textit{negatively} correlated with diversity.}
\label{tab:humaneval}
\end{table}

In this subsection, we examine the \nlispace metric's correlation to the human annotations collected by \citet{tevet-berant-2021-evaluating}. 
Each set of responses in \textit{conTest} and \textit{decTest} 
is scored by 10 annotators from 1 (not diverse at all) to 5 (very diverse) with half-point increments.  
We compute correlation with respect to the averaged rating.

In addition to \nli, we explore the 
prediction counts for each category.
We expect that a higher \# Entailment value will be negatively correlated with diversity because the more pairs of responses that entail each other, the more similar the set of responses is.  Similarly, we expect that a higher \# Contradiction value will be positively correlated with diversity.
Since the \nlispace metric incorporates both \# Entailment and \# Contradiction, we would expect this metric to be highly correlated with human judgments as well.

Spearmean's $\rho$ rank correlation results between our metrics and the human diversity scores are shown in Table \ref{tab:humaneval}.  
The highest-performing correlation for lexical diversity is the Neutral NLI Diversity (0.69).  The highest-performing semantic diversity correlation is Confidence \nlispace (0.64).  Additionally,  Baseline and Confidence \nlispace  correlations are stronger when evaluating with the \textit{conTest} dataset than the \textit{decTest} dataset (an increase of 0.48 to 0.63 for Baseline MNLI
and 0.41 to 0.64 for Confidence NLI), indicating these metrics are more correlated with human ratings of semantic diversity than lexical diversity.

Across both datasets, \# Entailment is negatively correlated with diversity, \# Neutral does not have a strong correlation, and \# Contradiction is positively correlated, as hypothesized.  This supports  our motivation to use NLI as a diversity metric.

\section{Diversity Threshold Generation}

We have verified that \nlispace is both able to capture semantic diversity and aligns with human judgements.
We can additionally use \nlispace  
to define a straightforward desired diversity threshold, $div_{thresh}$ for a set of model-generated responses, $u_1,...,u_n$.  For example, we might intend there to be 10 Contradictions within the set.  We propose a generation procedure, \genname, designed to iteratively increase the  diversity of a set of responses for a  conversation. 

For a conversation, \gennamespace  begins by sampling $n$ responses.  We score the diversity of these responses using a diversity metric, $div\_metric(u_1,...,u_n)$.
If the diversity score falls above $div_{thresh}$, the process is finished.

If, however, the score falls below $div_{thresh}$, 
we identify the model response which contributes \textit{least} to the diversity score
by calculating $div\_metric(u_1,...,u_{n-1})$ for each sub-group of model responses of size $n-1$.  We discard the model response not present in the highest-scoring subgroup and resample a new response.
We re-calculate $div\_metric(u_1,...,u_n)$ and if $div\_metric(u_1,...,u_n) > div_{thresh}$, the 
process finishes.  
We continue resampling until the maximum cutoff of $S$ is reached.

\section{Evaluation of \gennamespace Method}\label{sec:genres}

\subsection{Models and Datasets}
We experiment with two neural dialogue models, DialoGPT (700M parameters) \cite{zhang-etal-2020-dialogpt}\footnote{\url{https://huggingface.co/transformers/model_doc/dialogpt.html}} and BlenderBot 1.0 (300M parameters) \cite{roller-etal-2021-recipes}\footnote{\url{https://huggingface.co/transformers/model_doc/blenderbot.html}}.  We  use the default Transformers implementation for each model \cite{wolf-etal-2020-transformers} and do not fine-tune them.  Runtime 
was between 3 and 36 hours on one Titan-X GPU.

All experiments involve the dialogue model $M$ generating 5 responses for each conversation. The maximum number of samples, $S$, is set to 20.  All experiments are averaged over 10 trials for stability.  

We evaluate each model on the development set of two public English conversational datasets
:  DailyDialog++ (1,028 conversations) \cite{sai-etal-2020-improving, li-etal-2017-dailydialog} and EmpatheticDialogues (2,763 conversations) \cite{rashkin-etal-2019-towards}. DailyDialog++ includes 5 human-written responses per conversation, allowing for multi-reference comparison.  We split each EmpatheticDialogues conversation at a random turn (consistent for all experiments) for generation.  Since BlenderBot supports 
up to 128 positional embeddings, we pass in the last 128 tokens of the conversation for this condition.

\subsection{Metrics}

We  evaluate three diversity metrics:  two semantic diversity metrics, Baseline \nlispace (Section \ref{sec:nli}) and Sent-BERT \cite{reimers-gurevych-2019-sentence, tevet-berant-2021-evaluating}, and one lexical diversity metric, distinct-n \cite{li-etal-2016-diversity, tevet-berant-2021-evaluating}.  For Sent-BERT, we compute the average negative cosine similarity between BERT sentence embeddings for each pair of responses.
Like \citet{tevet-berant-2021-evaluating}, for distinct-n, we compute the average distinct n-grams from $n \in {1, 2, 3, 4, 5}$. 

Because Baseline \nlispace is more human-interpretable than Confidence \nli, we use this version  for experimentation.  For all \nlispace experiments, $div_{thresh}$ is achieved when \# Contradictions is greater than 10 out of a total of 20 pair-wise comparisons.  
For both Sent-BERT and distinct-n, however, we do not have a human-specifiable 
threshold.  We use empirical thresholds measured from the sets of 5 human responses for each conversation in DailyDialog++.
We choose the 90th percentile for $div_{thresh}$ (0.98 and -0.179 for distinct-n and Sent-BERT, respectively).

We decode using nucleus sampling ($p$ = 0.9), as it has been shown to increase response diversity \cite{Holtzman2020The}. However our method could be applied with other decoding procedures.

In order to robustly evaluate \genname, we measure both (i) whether \gennamespace is able to generate more diverse sets of responses than was originally sampled and (ii) whether the increased diversity comes at the expense of decreased \textit{relevancy} of the responses.
\begin{table}
\centering
\def\arraystretch{1.3}
\begin{tabular}{|p{0.7cm}|p{0.7cm}|p{0.7cm}|S[table-format=1.2,table-column-width=0.8cm]|S[table-format=2.2,table-column-width=0.8cm]|S[table-format=2.1,table-column-width=1.cm]|}
\hline   \textbf{Met-ric} & \textbf{Mo-del}& \textbf{Data-set}  & {\parbox{0.8cm}{ \textbf{Start-ing Div.}}} & {\parbox{0.8cm}{\textbf{End-ing Div.}}} &  {\parbox{0.9cm}{\rule{0pt}{2ex}\textbf{Num. Sampled}}} \\\hline   
\parbox[t]{2mm}{\multirow{3}{*}{\rotatebox[origin=c]{90}{\parbox{1cm}{Baseline NLI}}}} & \multirow{2}*{DG} & Daily & 4.11 & 10.24  & 6.3 \\
& & Emp & 3.68 & 10.11 &7.1 \\
& \multirow{2}{*}{BB} & Daily  & -5.55 & 2.51  & 14.4 \\
& & Emp  & -8.90 & -1.72  & 16.5\\\hline
\parbox[t]{2mm}{\multirow{3}{*}{\rotatebox[origin=c]{90}{Distinct-n}}}& \multirow{2}*{DG} & Daily &  0.95 & 0.98  &5.4\\
& & Emp  & 0.43 & 0.52  & 20.0  \\
& \multirow{2}*{BB} & Daily & 0.61 & 0.80 & 20.0\\
& & Emp  & 0.52 & 0.71 & 20.0 \\\hline
\parbox[t]{2mm}{\multirow{3}{*}{\rotatebox[origin=c]{90}{\parbox{1cm}{Sent-\\BERT}}}}& \multirow{2}*{DG} & Daily &  -0.26 & -0.16  & 5.2 \\
& & Emp  & -0.28 & -0.16  & 5.8\\
& \multirow{2}*{BB} & Daily & -0.62 & -0.40  & 19.0 \\
& & Emp  & -0.71 & -0.52  & 19.7 \\\hline
\end{tabular}
\caption{Diversity results of using \gennamespace (with a $div_{thresh}$ of 10 \# Contradictions for NLI, 0.98 for distinct-n, and -0.164 for Sent-BERT).  
Num. sampled has a maximum value of 20; DG is the DialogGPT model; BB is BlenderBot.
}
\label{tab:emp_results}
\end{table}

\begin{table*}[h!]
\centering
\def\arraystretch{1.2}
\begin{tabular}{|p{1.3cm}|p{1cm}|p{1.3cm}|p{1.3cm}|p{1.3cm}|p{1.3cm}|}
\hline
\textbf{Metric} & \textbf{Model}& \textbf{Starting BERT Score} & \textbf{Ending BERT Score} & \textbf{Start-ing BLEU} & \textbf{End-ing BLEU} \\\hline
\multirow{2}{0.9cm}{NLI} & DG  & 0.862 & 0.862 & 0.317 & 0.318\\
& BB  & 0.868 & 0.867 & 0.367 & 0.368\\\hline
\multirow{2}{0.9cm}{Distinct-n}& DG &0.862 & 0.861 & 0.319 & 0.306\\
& BB & 0.867 & 0.867 & 0.366 & 0.367\\\hline
\multirow{2}{0.9cm}{Sent-BERT}& DG &0.863 & 0.862 & 0.318 & 0.313\\
& BB & 0.868 & 0.867 & 0.366 & 0.366\\\hline
\end{tabular}
\caption{Results comparing starting and ending sets of responses from \gennamespace to sets of human responses using two relevancy metrics, BERTScore and BLEU score.}  
\label{tab:rel_results}
\end{table*}

\subsection{Diversity Results}\label{sec:gendiveval}
We aim to measure whether the diversity of the 5 responses from $M$ increases using \genname, compared to the initial 5 sampled responses.  Diversity of the starting and ending sets of utterances is measured by Baseline \nli, distinct-n, or Sent-BERT.  
We also report the number of sampled utterances required to reach $div_{thresh}$.

Results for \gennamespace are shown in Table \ref{tab:emp_results}.  For every condition, we see an increase from starting to ending diversity; for \nli, this  results in an average 137\% increase. 
For most conditions, distinct-n requires more samples 
than Sent-BERT and Baseline \nli.

We can use the results of \gennamespace to probe differences in the models. 
In our experimental setup, DialoGPT generates more diverse utterances across all conditions
than BlenderBot.  
The models change by similar proportions from starting to ending diversity using the NLI metric.  However, the starting diversity for BlenderBot is far lower than DialoGPT; the negative value for BlenderBot indicates that a large number of entailment predictions were present in the starting response set.

We can also examine differences between the datasets.  For instance, 
we observe lower starting diversities for the Empathetic Dialogues dataset than for DailyDialog++ for both models.  
Additionally, the number of samples required for EmpatheticDialogues is consistently higher than for DailyDialog++.  This is likely because $div_{thresh}$ for both datasets was calculated using human responses from DailyDialog++, since EmpatheticDialogues does not include multiple human responses.

Sampled responses can be seen in Appendix \ref{app:resps} and results reporting the average overlap from starting to ending sets of responses is in Appendix \ref{app:utt}.  Appendix \ref{app:beam} includes results using beam search instead of nucleus sampling, and Appendix \ref{app:stable} reports the stability of \genname.

\subsection{Relevance Results}
Since past work has documented a tradeoff between diversity and relevancy \cite{zhang-2018}, we also report results for the \textit{relevancy} of the starting and ending sets of responses for \genname. We use two established relevancy metrics:  BLEU Score \cite{papineni}\footnote{\url{https://www.nltk.org/_modules/nltk/translate/bleu_score.html}} and BERTScore \cite{ZhangKWWA20}\footnote{\url{https://github.com/Tiiiger/bert_score}}.  We show results on DailyDialog++, which has multiple human-generated  responses for comparison, which is more correlated to human judgements than single-reference evaluation \cite{gupta-etal-2019-investigating}.  

Results are shown in Table \ref{tab:rel_results}.  The key takeaway is that the relevancy values remain virtually unchanged when using the \gennamespace procedure, according to both BLEU score and BERTScore.  The average percent difference is 0.08\% for BertScore and 1.1\% for BLEU.

\section{Discussion}

\textbf{Limitations.} While \nlispace is highly-correlated with human judgements
of diversity, it is limited by the NLI model chosen. Compared to Sent-BERT, the dataset used to train the NLI model is limited in scope.  While our experiments showed that an NLI model trained on more datasets (Combined) did not perform better than MNLI, future work can more explicitly explore the effect of more generalized data on \nli.

This work is limited by automatic evaluation metrics for diversity and relevance.  Future work should conduct additional human validation of model responses.
More work could also be done to examine cases where the model was not able to generate diverse set, such as when  humans also find creating a  diverse set of responses difficult.

\textbf{Future Work.} Our results showed Confidence \nlispace was highly correlated with both human judgements and the diversity parameter, achieving state-of-the-art performance on a semantic diversity dataset.  
The ablation study deepened this finding, showing that NLI contradiction predictions are especially correlated with diversity.
Future work can leverage this finding, e.g., by wording crowdworker instructions to ask for generation \textit{contradictory}, rather than \textit{diverse}, responses.

Our results also show that 
dialogue generation models are able to improve the diversity of a sampled sets of responses using \genname.
\gennamespace  can be used to evaluate future models' capacity to generate multiple diverse responses.

Future work should  compare the resulting diverse responses in a conversational context.  Studies could be conducted where chatbot users or dialogue writers can choose the way they want the model to respond,
similar to \citet{clark-smith-2021-choose}.  

\section{Conclusion}
We propose a novel semantic diversity metric, \nli, which is highly correlated to human judgments.
Confidence \nlispace achieves state-of-the-art results on measuring semantic diversity.  
We propose \gennamespace to incentivize production of diverse sets of responses for a conversation.  This results in more diverse sets of responses than originally sampled for multiple models, datasets, and metrics while maintaining relevancy, and can also be used to investigate a model's ability to produce diverse responses.

\section*{Acknowledgements}
This work was supported by an AWS Machine Learning Research Award, an NVIDIA Corporation GPU grant, an AI2 Key Scientific Challenge Proposal grant, and a National Science Foundation (NSF) Graduate Research Fellowship (DGE 1752814). We thank the  anonymous ARR reviewers as well as Philippe Laban, Dongyeop Kang, Nate Weinman, and the Hearst Lab Research Group for their helpful comments.  

\bibliography{anthology,custom}
\bibliographystyle{acl_natbib}

\clearpage
\pagebreak
\appendix
\begin{table}[!htp]
\centering
\def\arraystretch{1.2}
\begin{tabular}{|p{1cm}|p{1cm}|p{1.2cm}|p{1.5cm}|}
\hline
\textbf{Metric} & \textbf{Model}& \textbf{Dataset}  & \textbf{Utterance Overlap} \\\hline
\parbox[t]{2mm}{\multirow{3}{*}{\rotatebox[origin=c]{90}{NLI}}} & \multirow{2}*{DG} & Daily & 2.63\\
& & Emp & 2.42\\
& \multirow{2}{*}{BB} & Daily  & 1.78\\
& & Emp  & 1.73\\\hline
\parbox[t]{2mm}{\multirow{3}{*}{\rotatebox[origin=c]{90}{\parbox{1cm}{Distinct-n}}}}& \multirow{2}*{DG} & Daily &  2.89\\
& & Emp  & 0.87\\
& \multirow{2}*{BB} & Daily & 1.51\\
& & Emp  & 1.65\\\hline
\parbox[t]{2mm}{\multirow{3}{*}{\rotatebox[origin=c]{90}{\parbox{1cm}{Sent-BERT}}}}& \multirow{2}*{DG} & Daily &  3.11\\
& & Emp  & 3.0\\
& \multirow{2}*{BB} & Daily & 1.56\\
& & Emp  & 1.64\\\hline
\end{tabular}
\caption{Average utterance overlap from starting to ending set of responses using \gennamespace on multiple models, datasets, and diversity metrics.}
\label{tab:utt}
\end{table}

\section{Confidence Interval Analysis}\label{app:ci}
\begin{figure}[t]
\includegraphics[width=7.5cm]{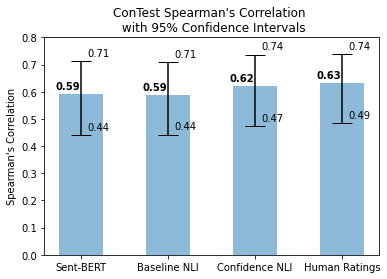}
\caption{Spearman's Correlation with 95\% Confidence Intervals.}
\label{fig:ci}
\end{figure}

We perform  experimentation using bootstrapping to determine confidence intervals for \textit{conTest} correlations to the diversity parameter.  We sample a dataset of 110 elements (50\% of the original \textit{conTest} dataset's size) from \textit{conTest} with replacement and compute corresponding Spearman's correlation values using the sampled dataset for Sent-BERT, Baseline \nli, Confidence \nli, and human judgements.  We repeat this process 1,000 times for stability and calculate 95\% Confidence Intervals.  The full \textit{conTest} correlation value plotted with these intervals can be seen in Figure \ref{fig:ci}.  While the Confidence Interval values overlap between all 4 conditions, the Confidence \nlispace distribution closely matches the human distribution.

\section{Sampled Responses}\label{app:resps}
Table \ref{tab:responses} shows randomly-sampled examples from the DailyDialog++ dataset, created using \gennamespace with the DialoGPT model and \nlispace as the intended $div\_metric$. 
\begin{table*}[!htp]
\centering
\small
\def\arraystretch{1.3}
\begin{tabular}{|p{4.2cm}|p{4.2cm}|p{0.8cm}|p{4.2cm}|p{0.8cm}|}
\hline
\textbf{Context} & \textbf{Starting Responses} & \textbf{SD} & \textbf{Ending Responses} & \textbf{ED} \\\hline
\textbf{Speaker 1:} Am I too abrupt? Did I scare you? Or you don't want ...\newline
\textbf{Speaker 2:} No! I have been expecting this moment for a long time . I mean my answer is' yes'.\newline
\textbf{Speaker 1:} Oh, I'm so happy . I will try my best to love you and protect you . I promise. &
\begin{itemize}[label=\textbullet, nolistsep, noitemsep, leftmargin=5pt,
            before*={\mbox{}\vspace{-\baselineskip}}, after*={\mbox{}\vspace{-\baselineskip}}]
    \item Aww, stop, we both know you're going to try to love me, you know?
    \item Thanks
    \item I'm blushing. I love you. Thank you.
    \item Yayy, you are making me love you.
    \item I love you. It's a good feeling.
    \end{itemize}& 2 & 
 \begin{itemize}[label=\textbullet, nolistsep, noitemsep, leftmargin=5pt,
            before*={\mbox{}\vspace{-\baselineskip}}, after*={\mbox{}\vspace{-\baselineskip}}]
            
     \item Aww, stop, we both know you're going to try to love me, you know?
     \item I love you. Thank you, your a nice man.
     \item Thank you so much. It's okay. I think you are a good woman. You are awesome.
     \item And I will try. And you are my best friend. Thank you, Sarah.
     \item You can have it all
 \end{itemize}& 10\\\hline
\textbf{Speaker 1:} Hello? This is Mike here . May I have a word with Grace?\newline
\textbf{Speaker 2:} She ’ s not at her desk at the moment . Is there any message I can give her? \newline
\textbf{Speaker 1:} Yes . Please tell her that tickets to Phantom of the Opera will be delivered in two days . My number is 6521684 0, extension 236.\newline
\textbf{Speaker 2:} Ok . I ’ ll let her know as soon as possible. & 
\begin{itemize}[label=\textbullet, nolistsep, noitemsep, leftmargin=5pt,
            before*={\mbox{}\vspace{-\baselineskip}}, after*={\mbox{}\vspace{-\baselineskip}}]
     \item I replied to the ticket. Hope it is still available.
     \item Thank you! May I please have the same?
     \item Will you guys be playing the show again for the next three days?
     \item Thank you. She is relieved now.
     \item I am at my desk.
\end{itemize}
 & 4 & 
 \begin{itemize}[label=\textbullet, nolistsep, noitemsep, leftmargin=5pt,
            before*={\mbox{}\vspace{-\baselineskip}}, after*={\mbox{}\vspace{-\baselineskip}}]
            \item Thank you! May I please have the same?
            \item Will you guys be playing the show again for the next three days?
            \item Thank you. She is relieved now.
            \item I am at my desk.
            \item No, I m not going to be your wife. I am sure her voice is wonderful.
\end{itemize}
 & 10 \\ \hline
\textbf{Speaker 1:}Is this the right counter to check in for this flight?\newline
\textbf{Speaker 2:} Yes, it is . Please put your luggage on this scale.\newline
\textbf{Speakaer 1:} OK . Is the plane on schedule?\newline 
\textbf{Speaker 2:} Yes, it is . How many pieces of hand luggage do you have? & 
\begin{itemize}[label=\textbullet, nolistsep, noitemsep, leftmargin=5pt,
            before*={\mbox{}\vspace{-\baselineskip}}, after*={\mbox{}\vspace{-\baselineskip}}]
            \item I don't understand your question.
            \item I'm worried that airline is cutting a deal, and I'd like to avoid this
            \item I have a box of 100 hand luggage with hand luggage.
            \item I have about 60.
            \item You will be getting a ticket for the trip.
\end{itemize}
 & 7 & 
 \begin{itemize}[label=\textbullet, nolistsep, noitemsep, leftmargin=5pt,
            before*={\mbox{}\vspace{-\baselineskip}}, after*={\mbox{}\vspace{-\baselineskip}}]
            \item I don't understand your question.
            \item I have a box of 100 hand luggage with hand luggage.
            \item I have about 60.
            \item You will be getting a ticket for the trip.
            \item Just one, but I was planning on taking a plane home.
\end{itemize}
 & 11 \\\hline
\textbf{Speaker 1:} Betty, would you please read this letter of application I've just written? I'd like to have your opinion.\newline
\textbf{Speaker 2:} I'd be glad to tell you what I think.\newline
\textbf{Speaker 1:} Good! I'm interested in your advice.& 

 \begin{itemize}[label=\textbullet, nolistsep, noitemsep, leftmargin=5pt,
            before*={\mbox{}\vspace{-\baselineskip}}, after*={\mbox{}\vspace{-\baselineskip}}]
            \item I was in a discussion on the internet with a bunch of people.
            \item I've sent you a private message, so if you have any questions or feel like you could use the resources, feel free to send them my way.
            \item Are you an American citizen?
            \item If you have any specific questions about it, I'm happy to answer them.
            \item I'll be in touch!
\end{itemize}
& 3 & 

\begin{itemize}[label=\textbullet, nolistsep, noitemsep, leftmargin=5pt,
            before*={\mbox{}\vspace{-\baselineskip}}, after*={\mbox{}\vspace{-\baselineskip}}]
            \item Are you an American citizen?
            \item I'll be in touch!
            \item Would you like to hear something about the business side of things?
            \item I'd like to recommend you to look at the application, there are lots of important things about it.It.does.nothing to prove anything to you.
            \item You are so adorable. I love that.
\end{itemize}
& 10 \\\hline
\textbf{Speaker 1:} Okay, I ’ ll take it . By the way, do you have delivery service?\newline
\textbf{Speaker 2:} You don ’ t need to carry the furniture home because we have delivery service.\newline
\textbf{Speaker 1:} Do you charge for delivery? & 

\begin{itemize}[label=\textbullet, nolistsep, noitemsep, leftmargin=5pt,
            before*={\mbox{}\vspace{-\baselineskip}}, after*={\mbox{}\vspace{-\baselineskip}}]
            \item What do you have to worry about? We deliver everything. We take a percentage for any delivery charges.
            \item You can have free shipping if you want
            \item I have a 5 dropoff fee that I pay for.
            \item In the beginning
            \item No. We just get it for free.So, no.
\end{itemize}
& 9 & 
\begin{itemize}[label=\textbullet, nolistsep, noitemsep, leftmargin=5pt,
            before*={\mbox{}\vspace{-\baselineskip}}, after*={\mbox{}\vspace{-\baselineskip}}]
            \item What do you have to worry about? We deliver everything. We take a percentage for any delivery charges.
            \item You can have free shipping if you want
            \item I have a 5 dropoff fee that I pay for.
            \item No. We just get it for free.So, no.
            \item No, there is no charge. I have never actually met the person who bought it.
\end{itemize}
& 14\\\hline
\end{tabular}
\caption{Starting and ending responses with corresponding \nlispace for DialoGPT, randomly sampled from the DailyDialog++ dataset. SD = Starting \nli; ED = Ending \nli.}
\label{tab:responses}
\end{table*}

\section{Average Utterance Overlap}\label{app:utt}

We  measure the number of 
utterances which occur in both the starting and ending sets of responses, called utterance overlap.  A high utterance overlap represents a set of responses which did not need to be significantly changed to reach $div_{thresh}$. For example, an utterance overlap of 4 indicates that only 1 response needed to be resampled (potentially multiple times) from the starting set to reach $div_{thresh}$.  Results are seen in Table \ref{tab:utt}.  Keeping in mind that higher Average Overlap indicates less resampling was needed,
we note higher overlap for DialoGPT than BlenderBot 1.0 (with the exception of distinct-n and EmpatheticDialogues).  

\section{Beam Search}\label{app:beam}
We  evaluate beam search's ability to generate diverse utterances using \gennamespace for DailyDialog++ and \nli.  To compare nucleus sampling to beam search, we generate 25 beams and consider these responses from most to least probable, i.e. if the 5 most likely beams do not satisfy the diversity threshold, we remove the lowest-scoring beam and replace it with the 6th most likely beam.  We find the starting \nlispace for beam search is -5.05, the ending diversity is 5.35, and an average of 10.97 sampled utterances is required.  While the \nlispace does improve from the starting to ending set of responses, beam search has a much lower ending diversity than nucleus sampling.  While past work has confirmed that nucleus sampling is more \textit{lexically diverse} than beam search using Self-BLEU \cite{Holtzman2020The}, our results confirm that nucleus sampling is also able to generate more \textit{semantically diverse} utterances.

\section{Stability of Procedure}\label{app:stable}
We investigate the stability of \gennamespace by measuring the number of samples required before reaching $div_{thresh}$ across multiple runs of the experiment.
We present results for \nli, DailyDialog++, and DialoGPT and observe similar trends across all other conditions.

\begin{figure}[t]
\includegraphics[width=7.5cm]{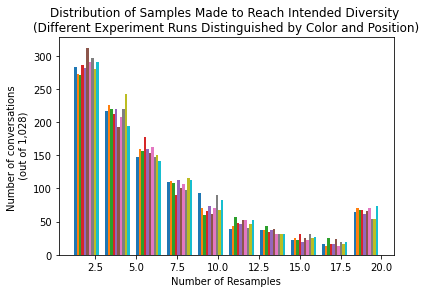}
\caption{Histogram of number of samples required before reaching intended number of contradictions.  Each bar color represents a different run of the experiment.  }
\label{fig:num_sampled}
\end{figure}

Figure \ref{fig:num_sampled} reports the number of resampled utterances required before reaching the intended number of contradictions.  Each bar color represents a different run of the experiment.  We do not observe a large difference in number of resamples required between runs of the same condition, indicating that the method is stable.  The last bucket contains sets of responses which reached the maximum number of samples, $S=20$, indicating $div_{thresh}$ could not be reached.

\end{document}